\documentclass[letterpaper]{article}
\usepackage[preprint]{css_2024}
\usepackage{multirow}

\usepackage{times}

\usepackage{mathtools}
\usepackage{graphicx}
\usepackage{mathrsfs}
\usepackage{amsmath}
\usepackage{amssymb}
\usepackage{natbib}
\usepackage{soul,color}
\usepackage{algorithm,algcompatible}
\usepackage{tabularx}
\usepackage{mathrsfs}
\usepackage{multirow}
\usepackage{bbm}
\usepackage{amsthm}
\usepackage{bbold}
\usepackage{color,soul}

\title{Uncovering Customer Issues through Topological Natural Language Analysis}

\author{ {\bf Shu-Ting~Pi} \\
Amazon\\
Cupertino, CA 95014\\
shutingp@amazon.com\\
\And
{\bf Sidarth Srinivasan}
\thanks{The work was carried out while SS was an intern at Amazon.}\\
Amazon           \\
Seattle, WA 98109 \\
sidarthsrini@g.ucla.edu\\
\And
{\bf Yuying Zhu}   \\
Amazon \\
Seattle, WA 98109    \\
imyuying@amazon.com
\And
{\bf Michael Yang}  \\
Amazon           \\
Seattle, WA 98109 \\
abyang@amazon.com
\And
{\bf Qun Liu}  \\
Amazon           \\
Seattle, WA 98109 \\
qunliu@amazon.com
}

\begin{document}

\maketitle
\begin{abstract}
E-commerce companies deal with a high volume of customer service requests daily. While a simple annotation system is often used to summarize the topics of customer contacts, thoroughly exploring each specific issue can be challenging. This presents a critical concern, especially during an emerging outbreak where companies must quickly identify and address specific issues. To tackle this challenge, we propose a novel machine learning algorithm that leverages natural language techniques and topological data analysis to monitor emerging and trending customer issues. Our approach involves an end-to-end deep learning framework that simultaneously tags the primary question sentence of each customer's transcript and generates sentence embedding vectors. We then whiten the embedding vectors and use them to construct an undirected graph. From there, we define trending and emerging issues based on the topological properties of each transcript. We have validated our results through various methods and found that they are highly consistent with news sources.
\end{abstract}

\section{Introduction}
E-commerce websites handle a vast number of online customer service requests daily. During a typical online customer service interaction, customers first interact with a chatbot which asks them questions to identify their intent. This intent is usually classified based on the product or service that the customer needs assistance with. For instance, an online consumer electronics retailer might use its chatbot to classify requests as relating to cell phones, computers, or home appliances, among others. The chatbot then routes the customer to an agent who specializes in the requested product or service to assist. While the actual business practices among companies may differ, the interaction process between customers and agents is generally similar. Agents usually begin with a greeting and ask for details about the customer's questions. They then engage in diagnosis and finally conclude with some closing remarks.

Generally, processing customer requests can take several minutes, making it one of the most time-consuming aspects of e-commerce business. Therefore, developing a standardized process to handle specific issues is critical to help customers save considerable time and optimize the available resources of agents. This is especially important during emerging events or sudden surges in customer inquiries. By anticipating common issues and developing standardized procedures for agents, businesses can improve their response times, reduce customer frustration, and ultimately enhance customer satisfaction. 
 
We present a novel machine learning framework, as illustrated in Fig.1, that can detect emerging and trending issues without predefined lists. \textbf{The terms "trending" and "emerging" refer to the topics that are most frequently discussed within the current time window and the topics that show the most rapid increase in discussion compared to the previous time window}. Our approach comprises three distinct components. Firstly, a deep learning model is employed, which utilizes an attention mechanism to automatically tag the primary question sentence in each customer's transcript and generates sentence-level embedding vectors. To improve the performance of cosine similarity, we decouple the covariance matrix to whiten the embedding vectors, bringing the coordinates of the feature space close to an orthonormal basis. We then construct an undirected graph based on cosine similarity. Finally, we analyze the topology of the graph by calculating the centrality of each customer's question. This enables us to quantify both trending and emerging issues.


\begin{figure*}
  \centering
  \includegraphics[width=0.9\textwidth]{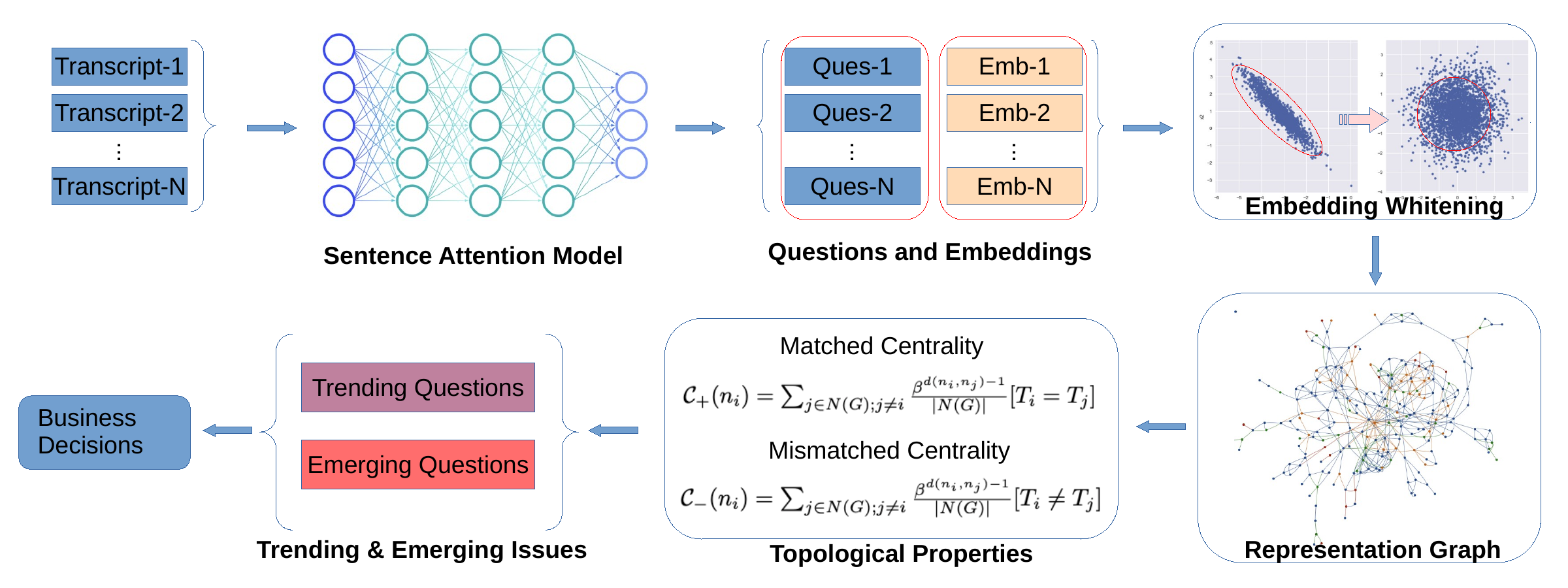}
  \caption{Our proposed workflow involves several steps. Initially, the transcripts are passed to a sentence attention model to extract the primary questions asked by the customers and their corresponding sentence embeddings. The embeddings are then whitened to obtain representations in an isotropic coordinate system. These whitened vectors are then utilized to construct an undirected graph, and their topological properties are calculated to identify both trending and emerging issues.}\label{workflow}
\end{figure*}  

\textbf{Related Works} Our work relates to several areas in the literature, including sentence-level attention, sentence tagging, and graph-based clustering. Works related to sentence attention include \citep{sentence-attention1}, \citep{sentence-attention2}, and \citep{sentence-attention3}, which apply the idea to document classification, summarization, and text noise reduction, respectively. Regarding sentence tagging, \citep{sent-tagging1} and \citep{sent-tagging2} are works that perform sentence tagging from token-level representation, and have also influenced our approach. Additionally, we reference several works in graph-based clustering, including \citep{graph-clustering1} and \citep{graph-clustering2}, which use this technique to handle repetitive sequences and multiview data. These works have inspired our research in the application of topological data analysis to natural language processing.

\section{Question Tagging and Sentence Attention Model} 
We present a deep learning model that can automatically tag the primary question in a contact transcript. This model is a crucial component for detecting both trending and emerging issues, as the identified questions will be utilized in later sections. 

\subsection{The Dataset}
There has been a lack of publicly available datasets related to customer service transcripts. To address this gap, we partnered with customer service team to initiate this research, using a dataset from an online chat system that enables customers to communicate with customer service agents. We collected over 500,000 contact transcripts during 2022, recording conversations between customers and agents. Each contact also comes with a unique label of the product or service that the customer and agents discussed. It's worth noting that the dataset only contains customer text data, with all confidential information, such as names and account details, anonymized to protect privacy before being shared with researchers. Although the dataset is from a specific database, the methodology presented in this article can be applied to other use cases as well.

\subsection{Sentence Embedding}
Our goal is to identify the primary question sentence in a customer-agent contact transcript. We propose two hypotheses: (1) the primary question sentence typically appears in the first few sentences of the customer's interaction with the agent, and (2) it contains the most relevant information about the product or service being discussed. If these hypotheses hold, we can treat the problem as a machine learning task: \textbf{identifying the customer sentences near the agent's initial response that are most useful in predicting the product or service of the contact for a machine learning classifier}. To achieve this, we need information on attention weights at the sentence level.

We propose a deep learning model, as shown in Fig. 2, to achieve our goal. Unlike traditional text classification models that represent an article as a 2D tensor $\in R^{\mathbb{N}_{at} \times \mathbb{N}_{we}}$, where $\mathbb{N}_{at}$ is the number of tokens in the article and $\mathbb{N}_{we}$ is the dimension of the word embedding, our approach represents each article as a 3D tensor $\in R^{\mathbb{N}_{as} \times \mathbb{N}_{st} \times \mathbb{N}_{we}}$. Here, $\mathbb{N}_{as}$ is the number of sentences in the article, and $\mathbb{N}_{st}$ is the number of tokens per sentence. To accommodate varying numbers of sentences and tokens per sentence in each transcript, we use zero-padding to ensure a consistent tensor shape for subsequent processing.

To obtain sentence-level embeddings, we treat each sentence as a temporal slice and apply a time-distributed wrapper to a sequence model $\Sigma$, such as BERT or LSTM. This ensures that the model receives only one sentence per time step, allowing us to embed each sentence. The resulting output tensor, $Q^{'}\in R^{\mathbb{N}_{as} \times \mathbb{N}_{se}}$, contains $\mathbb{N}_{as}$ sentence vectors, each with $\mathbb{N}_{se}$ dimensions.

Our "bag of sentences" model currently does not consider sentence positions, but we have observed that customer questions tend to appear in early sentences during interactions with agents. This suggests that sentence positions can impact attention weights, so we incorporate sentence position information into the model.

To do this, we adapt the idea of position embedding used in many language models for tokens, but apply it to sentences. We assign each sentence an index, ranging from -$\mathbb{N}_{as}$ to +$\mathbb{N}_{as}$, representing the number of sentences between the current sentence and the agent's first response. For instance, an index of -5 indicates that the sentence is five steps before the agent's first sentence, while +5 indicates five steps after. We shift the indices by $\mathbb{N}_{as}$, resulting in an allowed index range of 0 to +$2\mathbb{N}_{as}$, with the sentence having an index of $\mathbb{N}_{as}$ being the agent's first sentence to avoid negative indices. For a sentence with index $i$, the $p$-th component of the position embedding vector $E_i$ is given by $E_i(2p)$ and $E_i(2p+1)$:

\begin{align}
E_i(2p) &= sin(i/10000^{2p/d_{pos}})\\    
E_i(2p+1) &= cos(i/10000^{2p/d_{pos}})
\end{align}
where $d_{pos}$ is the dimension of embedding vector. By adding $Q^{'}_{i}$ and $E_{i}$, we get the final sentence embedding vector $Q_{i} = Q^{'}_{i} + E_{i}$.

\begin{figure*}
  \centering
  \includegraphics[width=0.8\textwidth]{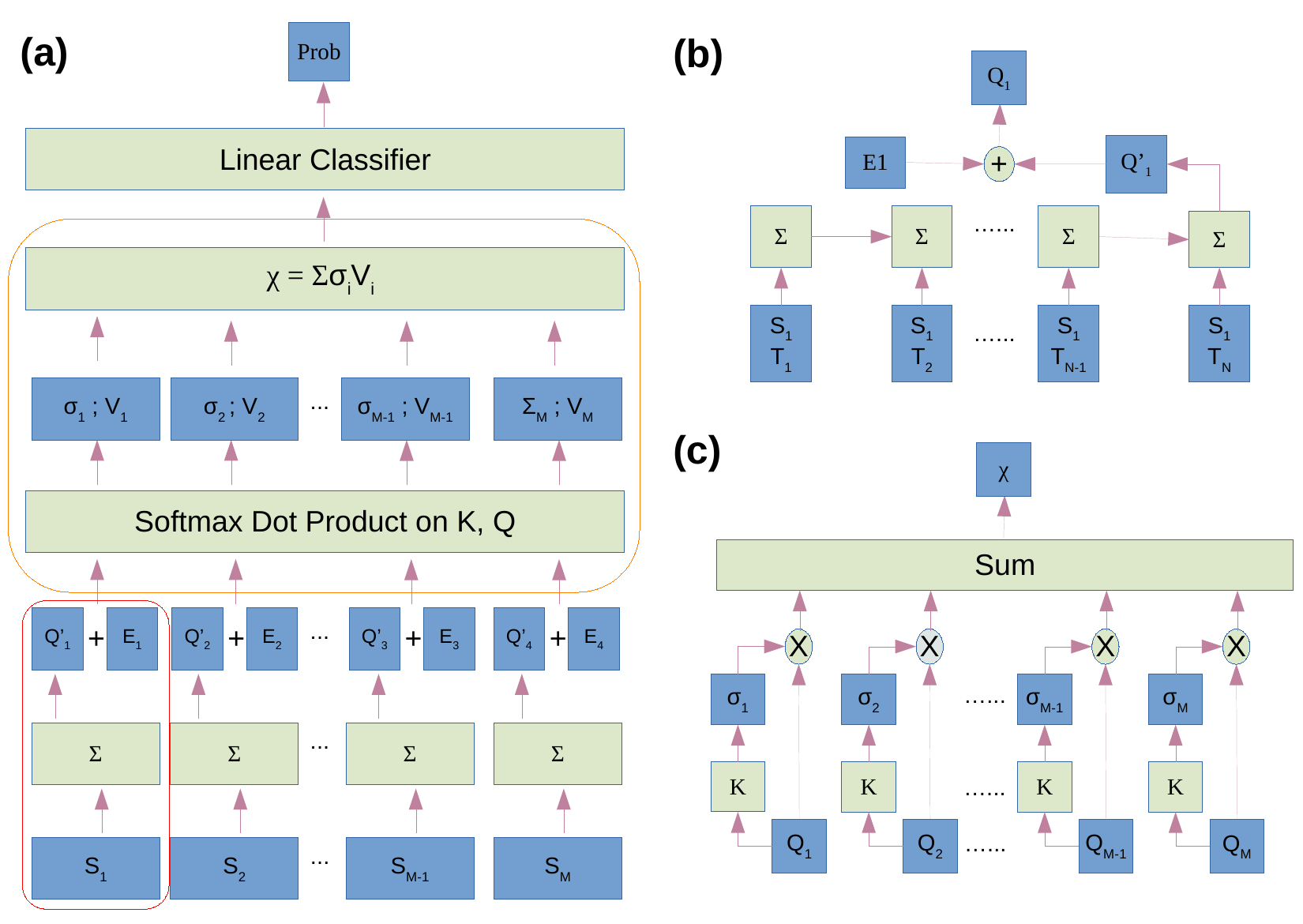}
  \caption{The Sentence Attention Model. The model consists of blue blocks, representing tensors, and green blocks representing operators. (a) The neural network is comprised of sentence tensors, $S_{i}$, and a sequence model, $\Sigma$, which outputs $Q^{'}{i}$. The position embedding vector $E{i}$ is combined with $Q^{'}{i}$ to create the sentence embeddings $Q{i}$. Finally, a linear classifier predicts the product/service. (b) The red block in (a) is described in detail. Tensor notation ($S_1;T_n$) refers to the $n$-th token in sentence $S_1$. The sequence model $\Sigma$ processes each word in a sentence using a time-distributed wrapper to handle multiple sentences. (c) The orange block in (a) is explained. A dense layer, $K$, with a softmax activation function is applied to all sentence embedding vectors $Q_{i}$ (via a time-distributed wrapper) to calculate attention scores $\sigma_{i}$. Note that $Q_{i}$ is equivalent to $V_{i}$.}\label{SeaCat}
\end{figure*}

\subsection{Sentention Attention}
We define sentence level attention\citep{transformer}: 

\begin{equation} Attention(\overrightarrow{K},\widehat{Q},\widehat{V)} = \overrightarrow{\chi} = softmax( \frac{\widehat{Q} \cdot \overrightarrow{K}}{\sqrt{d_{k}}})^{T} \cdot \widehat{V} = \overrightarrow{\sigma}\cdot \widehat{V}     
\end{equation}

, where $\widehat{Q}\in R^{\mathbb{N}_{as}\times\mathbb{N}_{se}}$ is a query tensor (i.e. the sentence embedding), $\overrightarrow{K}\in R^{\mathbb{N}_{se}}$ is a key vector and  $\widehat{V}\in R^{\mathbb{N}_{as}\times\mathbb{N}_{se}}$ is a value tensor. We can generally define $\widehat{Q} = \widehat{V}$ without any impact on model performance. As a result, the attention value $\overrightarrow{\chi} \in R^{\mathbb{N}_{se}}$ is a vector and the sum of its elements must equal 1, due to the application of the softmax function. 

Our model's attention vector $\overrightarrow{\chi}$ has a simple interpretation. As shown in Fig. 2, each sentence in a transcript is represented as a vector $\overrightarrow{V}_i$ (the $i$-th row of the 2D tensor $\widehat{V}$). The attention vector $\overrightarrow{\chi}$ is a linear combination of all sentences, computed as $\overrightarrow{\chi} = \sum_{i}\sigma_{i} \overrightarrow{V}_i$, subject to the constraint that $\sum_{i}\sigma_{i}=1$. We then pass $\overrightarrow{\chi}$ through a fully connected layer with a softmax activation function to predict the product or service associated with the transcript. The attention weights $\sigma_{i}$ reflect the importance of each sentence in determining the product or service, with higher weights assigned to sentences containing more critical information.

\subsection{Experiments}
Our model was trained on 500,000 transcripts with 152 classes to predict the product or service discussed in each contact. We used DistilBERT \citep{distill_bert} as the embedding model $\Sigma$, with an output dimension of 768. To prepare the transcripts for training, we padded each one with zeros to create 64 sentences, each with 128 words.

After training, we calculated attention weights $\sigma_{i}$ for each customer sentence in a transcript. To identify the primary question, we assumed that it is the sentence with the highest attention weight that is $N$ steps before or after the agent's first sentence. In testing on 4,000 human-annotated transcripts, we found that $N=2$ (i.e., $\pm2$ steps) yielded the best results, correctly identifying the primary question sentence in 84.3\% of the total transcripts. Fig. 1 in the Appendix illustrates an example of how our sentence attention model tags the primary question sentence.

To evaluate the impact of sentence position embedding on performance, we compared the model with and without this feature. Although the difference in accuracy was minimal (83.4\%), we observed that sentence position embedding resulted in higher attention weights on the primary question sentence. Further investigation is needed to explore the effects of position embedding.   

\section{Topological Natural Language Analysis}
Our objective is to identify both emerging and trending topics among the questions gathered by the model presented in Sec. 2. However, conventional clustering methods face several challenges in achieving this goal, including sensitivity to hyperparameters, scalability issues with large numbers of classes and samples, and lack of flexibility in defining distances.

Moreover, identifying emerging topics involves changes in the volume of a topic between time windows, and conventional clustering methods cannot determine whether two clusters in different datasets are related to the same topics. It is even possible for an emerging topic to be present in the current time window without appearing in the previous time window. Additionally, conventional clustering methods may treat emerging topics as noise due to their much smaller volume than trending topics.

To overcome these challenges, we propose a topological-analysis-based method    robust to hyperparameter selection and can quantitatively detect both emerging and trending topics between different time windows.

\subsection{Sentence Embedding Whitening}
Assuming we have gathered a substantial number of customer's primary questions and their associated embedding vectors ${Q^{'i}_{t}}$, where $Q^{'}$ represents the output embedding vectors of $\Sigma$ without the addition of position embedding, $i=1\sim N$ corresponds to the $i$-th transcript, and $t$ signifies the tagged sentence. Provided that we can define a metric or distance to express the semantic similarity between two questions, it becomes possible to construct a graph in which an edge connects two questions if their cosine distance falls below a specified threshold.

However, cosine distance may not be an appropriate metric for measuring text similarity. In fact, several experiments have shown that cosine similarity is not suitable for use with BERT-based representations and its performance in many similarity tasks is inferior to that of traditional embedding methods\citep{poor-bert} such as GloVe\citep{glove} or Word2Vec\citep{word2vec}. This is mainly due to the fact that cosine distance assumes an orthonormal coordinate system, which is not the case for most pre-trained sequence models. Various approaches have been proposed to address this issue, such as BERT-flow\citep{bert-flow} or representation whitening\citep{whitening}, to ensure that data distributions tend to be isotropic, a property that an independent basis set should have. Representation whitening, a post-processing method that does not require model training, is particularly effective in calibrating sentence representations. Hence, we adopt this strategy to improve the performance of our sentence representation.

The core idea of representation whitening is to find a linear transformation that sets the mean of the set $\{Q^{'i=1\sim N}_{t}\}$ to zero and the covariance matrix to an identity matrix. To achieve this, we first calculate the mean $\mu$ and covariance matrix $K$. As we aim for a diagonalized covariance matrix, we can compute the unitary matrix $U$ and singular values matrix $A$ using singular value decomposition (SVD)\citep{svd}: $U,A,U^{\dagger} = SVD(K)$. In this case, the whitening matrix $W$ that diagonalizes the covariance matrix $K$ is:
$W^{\dagger}KW = I\quad; W = U\sqrt{A^{-1}}$.
Once we obtain the whitening matrix $W$, the whitened sentence embedding vectors become:
$z_{i} = (Q^{'i}_{t}-\mu)W$.
Under this transformation, the mean of $\{z_i,i=1\sim N\}$ equals zero, and the covariance matrix being equal to an identity matrix $I$ is guaranteed. Although we lack labels to evaluate the performance of whitened sentence embedding vectors in capturing similarity, we manually inspected a small portion of the data with cosine similarity. Whitened vectors $z_i$ indeed appear more reasonable than the original ones. From this point, we will use $z_i$ as the embedding vector of the tagged question sentence in the $i$-th transcript unless specified otherwise.

\subsection{Undirected Representation Graph and Centrality}
Let us assume that we have gathered primary customer questions and their corresponding sentence embeddings, ${z_i}$, over two time periods, $T_0$ and $T_1$. Each sentence can be represented as a node in an undirected graph, with edges connecting nodes whose cosine similarity surpasses a specified threshold, $\alpha$. It is important to note that this graph is constructed using data from both time periods, $T_0$ and $T_1$. In this graph, a node with a higher number of neighbors indicates a greater number of similar questions.

Centrality is a significant topological property of a node, which describes its importance within a graph. Various types of centrality exist, each applicable to different scenarios. In this case, we define two new types of centrality that modify the decay centrality. Assuming we have an undirected graph represented by an adjacency matrix $A$, where $A(i,j)=1$ if nodes $cos(z_i,z_j) \ge \alpha$, the decay centrality of a node $n_i$ in a graph $G$ is defined as\citep{centrality}:
\[\mathcal{C}(n_i) = \sum_{j \in N(G); j\neq i} \frac{\beta^{d(n_i, n_j)-1}}{|N(G)|}\]
Here, $N(G)$ refers to the set of nodes in the graph, and $|N(G)|$ is the total number of nodes in the graph. This normalization factor ensures that the centrality $\mathcal{C}(n_i)$ is independent of the graph's size. The attenuation factor $\beta$ is typically selected such that $0<\beta<1$. The topological distance $d$ between nodes $i$ and $j$ is the graph distance, not the Euclidean distance $|z_i-z_j|$. The numerator is given by an exponent $d(n_i, n_j)-1$, which results in $\beta^{d(n_i, n_j)-1}=1$ when nodes $i$ and $j$ are directly connected. A question with higher decay centrality means the graph has more similar questions. 

Given that time is an additional property of each node, we modify the definition above and propose two new types of centrality:
\begin{itemize}
    \item matched decay centrality:
        $\mathcal{C}_{+}(n_i) = \sum_{j \in N(G); j\neq i} \frac{\beta^{d(n_i, n_j)-1}}{|N(G)|}[\mathcal{W}_i=\mathcal{W}_j]$\\
    \item mismatched decay centrality:
        $\mathcal{C}_{-}(n_i) = \sum_{j \in N(G); j\neq i} \frac{\beta^{d(n_i, n_j)-1}}{|N(G)|}[W_i\neq W_j]$ \\
\end{itemize}
In these equations, the brackets $[\cdot\cdot\cdot]$ represent Iverson brackets, yielding 1 if the statement inside is true and 0 if false. $\mathcal{W}_i$ refers to the time window that node $i$ belongs to. $\mathcal{C}_+$ considers contributions only when two nodes belong to the same time window, while $\mathcal{C}_-$ accounts for contributions from different windows, decaying exponentially as the distance increases. It is important to note that node $j$ can reach node $i$ through intermediate nodes in either the same or different time windows. Furthermore, we have $\mathcal{C}(n_i) = C_{+}(n_i)+C_{-}(n_i)$. This definition splits the decay centrality into two terms based on the time window. We can effectively identify trending and emerging issues by utilizing matched and mismatched decay centrality. 

\begin{figure*}
  \centering
  \includegraphics[width=0.9\textwidth]{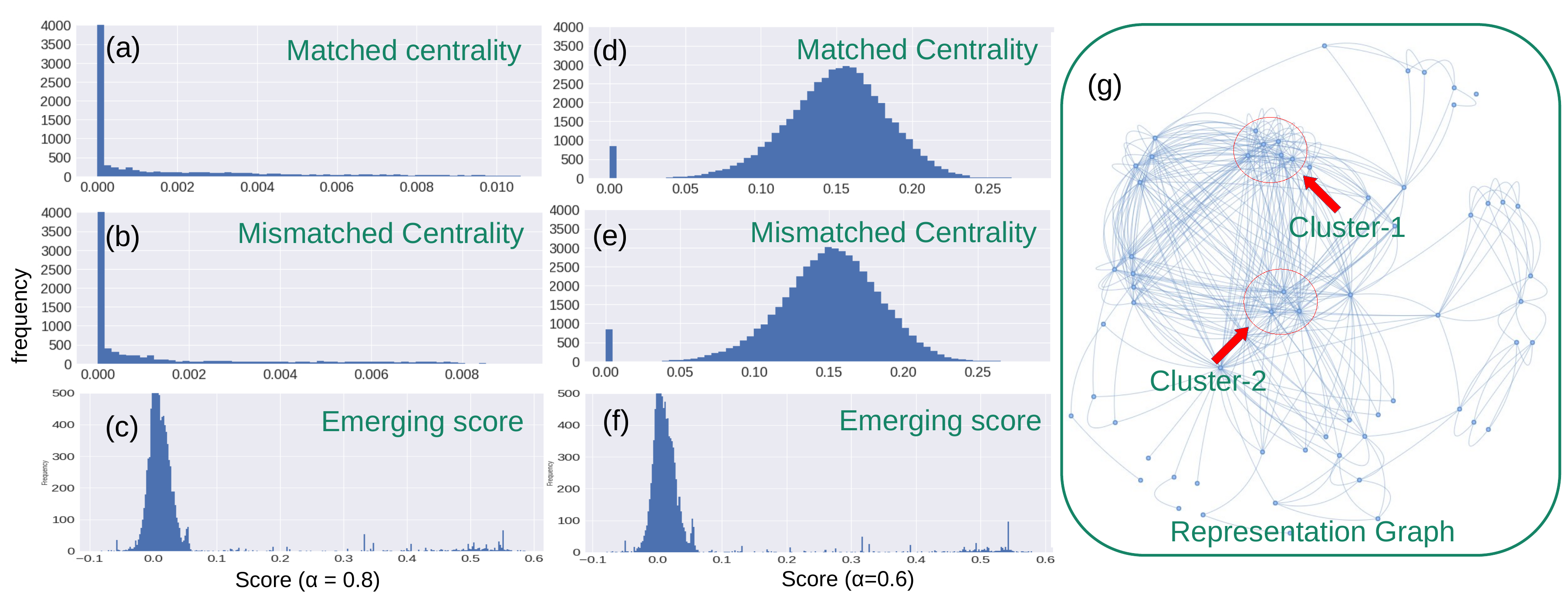}
  \caption{The centrality distribution of Fire Tablet in Feb 2023 is depicted in panels (a)-(f), where we compare the cosine similarity thresholds of $\alpha = 0.8$ and $\alpha = 0.6$. As for panel (g), the graph built using a few selected samples clearly demonstrates that similar sentences tend to cluster together, resulting in high centrality for nodes around the cluster centers. }\label{celtrality_dist}
\end{figure*}   
\section{Trending and emerging Issues}

\subsection{Topological Perspective}
We are only interested in the nodes in the current time window, so we define trending and emerging issues from a topological perspective as follows: 1). Trending issues are identified as the customer's primary questions with a large trending score:  $\mathcal{S}_t =  \mathcal{C}_{+}$ 2). Emerging issues are identified as the customer's primary questions with a large emerging score:  $\mathcal{S}_e = tanh^{2}(\mathcal{C}_{+}/\gamma)[\mathcal{C}_{+}-\mathcal{C}_{-}]/\mathcal{C}$.
The definitions for the trending and emerging scores are simple to understand. The trending score, denoted by $\mathcal{S}_t$, counts the number of similar questions in the graph that exist in the current time window. On the other hand, the emerging score, given by $[\mathcal{C}_{+}-\mathcal{C}_{-}]/\mathcal{C}$, represents the difference in centrality contributed from the previous time window and the current time window. It's important to note that $\mathcal{C}_++\mathcal{C}_-=\mathcal{C}$. As a result, the emerging score lies between -1 and +1, where -1 and +1 correspond to the centrality entirely coming from the previous and current time window respectively, while 0 indicates equal contributions from both time windows. To avoid identifying small-size clusters that may not be significant for a business that receives a large volume of customer contacts, we introduce an additional filtering factor, $tanh^2(\mathcal{C}_+/\gamma)$, where $\gamma$ is a parameter that controls the strength of the filter. This factor applies a weight to the emerging score, such that $tanh^2(\mathcal{C}_+/\gamma)$ drops to 0 when $\mathcal{C}_+\ll \gamma$, and saturates to 1 when $\mathcal{C}_+\gg \gamma$.

\subsection{Experiment}
We constructed a graph using the MessageUs transcript data by applying a cosine similarity threshold of $\alpha=0.7$, an attenuation factor of $\beta=0.5$, and a filter factor of $\gamma=0.1\times Max(\{ \mathcal{C}_{+}^{i}; i=1\sim N \})$, where $N$ is the total number of nodes. This filter factor was set to 10\% of the maximum number of $\mathcal{C}_{+}$ values, and it helped to exclude small clusters that were unlikely to be significant. We used topological data analysis to identify the most prominent clusters, and we found that the topics of these clusters were highly consistent, even though the exact values of $\mathcal{C}_+$ and $\mathcal{C}_-$ were sensitive to the chosen hyperparameters. Varying the hyperparameters affected only the sizes of the clusters and did not significantly alter the topics we discovered, which demonstrates the robustness of the topology analysis approach.

Fig. 3 shows the distribution of $\mathcal{C}_+$, $\mathcal{C}_-$, and $\mathcal{S}_e$ for Tablet-related transcripts, using Jan 2023 and Feb 2023 as the previous and current time windows, respectively (around 100K data points). We compare the distributions for a cosine similarity threshold of $\alpha=0.8$ (panels (a)-(c)) and $\alpha=0.6$ (panels (d)-(f)). Decreasing $\alpha$ results in a more compact graph and a more Gaussian-like distribution of centrality. We observe that the emerging score distribution can be separated into two parts: a Gaussian-like distribution around zero due to the filter factor applied to nodes with $\mathcal{C}_+\ll \gamma$, and a long-tail region due to nodes with $\mathcal{C}_+\gg \gamma$. The filter factor helps to focus on the outliers, and the nodes in the long-tail region are generally similar, with high-score nodes unlikely to become low-score nodes due to changes in hyperparameters.

It is important to note that similar sentences tend to form clusters. As a result, the neighbors of a high centrality node also tend to have high centrality, as shown in Fig. 3(g). To avoid locating the same cluster multiple times, it is necessary to ensure that two centers are sufficiently far apart. To achieve this, we first designate the node with the largest $S_t$ (for trending) or $S_e$ (for emerging) as the cluster center and its surrounding neighbors with graph distance $\leq 3$ as the cluster members. Next, we search for the node with the next-largest score at least a graph distance of 4 away from any known clusters as the next cluster center. We repeat this procedure until we have obtained the desired number of clusters.

Table.1 of the appendix provides examples of sentences from the top-1 trending and emerging clusters of Tablet in February 2023. As one can see, all the customer's questions within each cluster are very similar, demonstrating the effectiveness of our approach in cosine similarity and topology-based topic detection.

\section{Result and Discussion}

We have successfully developed a machine learning model capable of extracting trending and emerging issues from extensive transcripts. However, directly validating the precision rate of our model presents a challenge. To address this challenge, we have proposed a human-annotation-based method for validation. 

Our data collection spanned from November 2022 to February 2023 and involved more than 10 different products, including Kindle, Echo, Music, eBook, and Prime Video. During this period, our focus was on identifying the top-3 trending and emerging issues. We took care to exclude issues that appeared unreasonable due to quality of data or instances of fraud or policy abuse attacks. As a result of this process, we successfully identified 84 trending issues and 64 emerging issues.

To assess trending issues, each annotator is asked to select three keywords that best represent each issue. We then count the number of transcripts containing these keywords simultaneously. Remarkably, we discovered that all 84 issues, representing 100\% of the total, constitute a large portion, at least 10\%, of the overall volume of each product line. Furthermore, the volume remains consistently stable from month to month. Although it's challenging to determine if these are the largest issues, it's noteworthy that a significant portion of them pertains to return or refund-related matters, aligning closely with our business experiences.

Regarding emerging issues, we employ three methods to identify them. Firstly, we select the three most representative keywords for each issue and monitor changes in volume within transcripts containing these keywords. Additionally, we investigate whether the issue surfaces in the Amazon Digital and Device Forum for that month, where customers discuss Amazon's services, with a minimum of 10 replies ( not necessary in a single thread ). Lastly, we conduct a Google search to determine if these issues are covered in news media.

Our surveys revealed some insightful findings: approximately 90\% of emerging issues exhibit volume changes exceeding 30\% between two months, around 60\% of the issues are discussed in the Amazon Digital and Device Forum for that month, and roughly 15\% of the issues are covered in news media, typically related to live events or new product launches. The surveys support our model indeed captures the emerging issues very well. In Table 2 of the appendix, we provide a few examples of emerging topics found in Feb 2023 and our validation results, and most of the topics align well with the news sources.

\section{Conclusion}
In summary, we have presented a unique machine learning framework for extracting customers' trending and emerging issues. Our work starts with an attention-based deep learning model that tags customers' primary questions and generates corresponding sentence embeddings simultaneously. We then transform the sentence embeddings into an isotropic coordinate system using whitening techniques to improve the cosine similarity performance. Finally, we apply topological natural language analysis methods to analyze the centrality of each question, enabling us to identify trending and emerging issues.

Our work makes a significant contribution by demonstrating the application of a sentence-level attention mechanism in conversational transcripts, an area that has been understudied. We combine this mechanism with topological data analysis to extract useful information for a real-world problem. 

\bibliographystyle{apalike}
\bibliography{citation.bib}

\newpage
\onecolumn

\appendix
\section{Appendix: Example of Transcript}
\setcounter{figure}{0}
\begin{figure*}[!htb]
  \centering
  \includegraphics[width=0.9\textwidth]{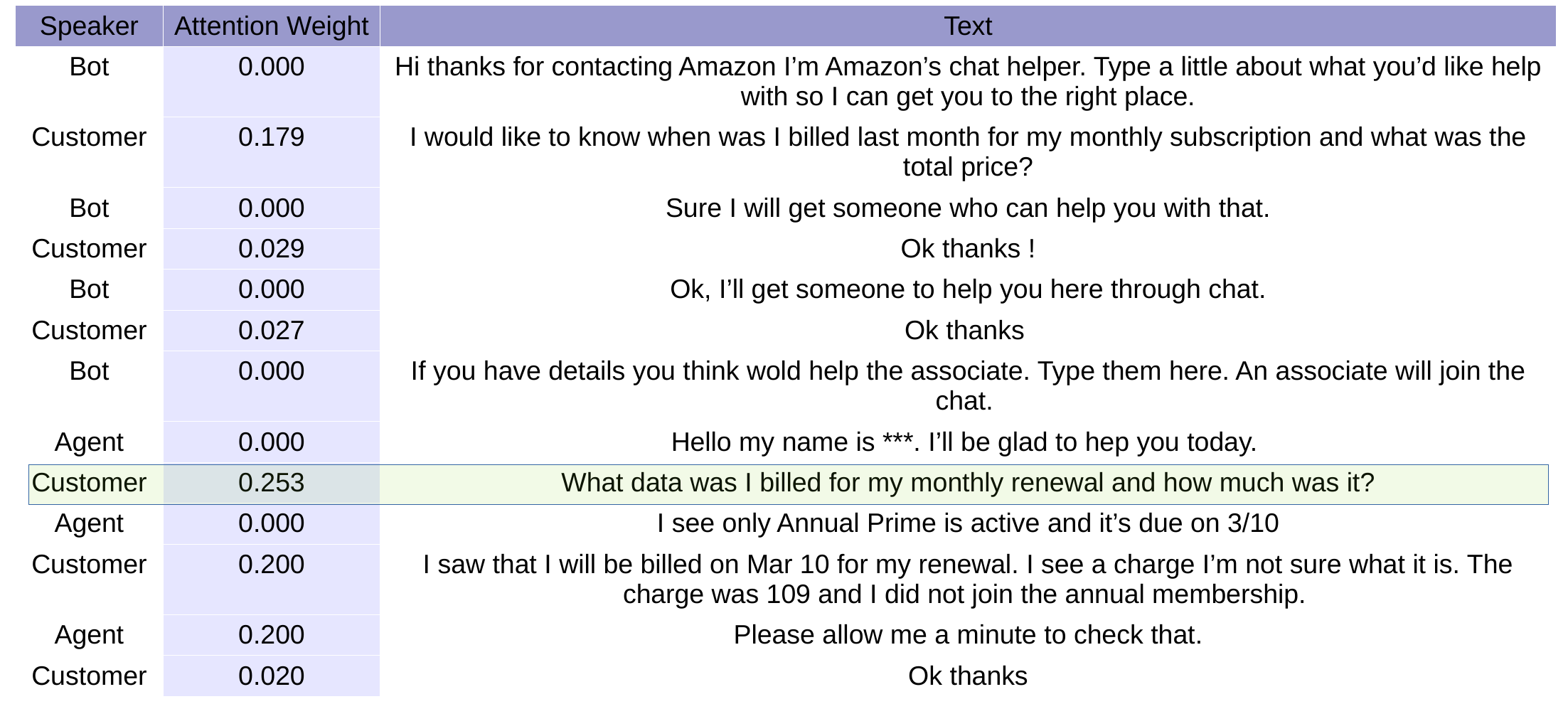}
  \caption{An illustration of the functioning of the question tagging model. The model calculates the attention score, $\sigma_{i}$, for each sentence in a transcript. The sentence preceding and following the agent's first sentence, with the highest score is predicted as the customer's primary question, i.e. the highlighted sentence.
 }\label{example}
\end{figure*} 

\section{Appendix: Trending \& Emerging Issues on Fire Tablet}
\begin{table*}[!hbt]
\centering
\begin{tabular}{| p{0.45\linewidth} | p{0.45\linewidth} |}
\hline
Trending Issues & Emerging Issues\\
\hline
\hline
I need to utilize the warranty for my kids' tablet with a cracked screen. & On the kids' tablet, whenever I try to access my child's profile, it consistently displays an error message saying "oops, something went wrong." However, when I switch to the parent profile, everything works fine.  \\
\hline\
I need to check the warranty for my son's cracked tablet screen. & Additionally, the kids' profile on the tablet consistently displays an error message saying "oops, something went wrong," while the adult profiles work fine. \\
\hline\
My son accidentally cracked the screen on our kid's tablet, and I'm looking to get it replaced under the attached warranty. & However, when I try to access my child's profile, I keep encountering an error message saying "oops, something went wrong."\\
\hline\
I would like to check if my tablet is still covered by warranty. & Furthermore, the kids' profile on the tablet is not functioning properly. Despite attempting a factory reset, I am consistently greeted with an error message stating "oops, something went wrong" whenever I try to load the kids' profile. However, the adult profile is functioning normally as usual.\\
\hline\
My daughter accidentally broke the screen of her tablet for kids. I was wondering if it's possible to file a claim to have it repaired. & Additionally, I'm experiencing difficulties accessing my child's page on the Tablet. While my information loads successfully, I keep encountering an error message saying "oops, something went wrong" when Kids profile begins to load.\\
\hline
\hline
\end{tabular}
\caption{In our analysis of Amazon Fire Tablet's top-1 trending and emerging clusters in Feb 2023, we found that customers reported cracked screens as a trending issue and sought warranty replacements or repairs, while problems with the kids' profile emerged as an emerging issue. The similarity among sentences in the trending and emerging topics extracted from the same cluster suggests the effectiveness of our method.}
\label{table1}
\end{table*}

\newpage
\section{Appendix: Selected Emerging Issues on Various Products}
\begin{table*}[!hbt]
\centering
\begin{tabular}{ | p{0.10\linewidth} || p{0.25\linewidth} | p{0.25\linewidth} | p{0.25\linewidth} |}
\hline
Product & Cluster Center Sentence & Topic Summarization & Validation\\
\hline
\hline
 Tablet & It keeps saying oops something went wrong when I try and get into my child's profile & Customers have reported encountering an error message when attempting to enter child's profile. & During February, this issue was extensively discussed in Amazon's Device Forum and was actively being addressed by Amazon's engineering team.  \\
\hline
 Music &  Hi I was getting an error code and I was unable to play any music from any play list & Customers have reported encountering an exception error with a specific code while playing music. & We compared the probability of finding the keywords "exception/play/error" together in a transcript and observed a 30\% increase from the previous month.  \\ 
\hline
Game \& Software &  Hi My preorder for the Hogwarts Legacy is VERY late It s supposed to arrive today but won't be shipped until 23 This is really unacceptable. & Customers have reported that their preordered game, Hogwarts Legacy, has been delayed. & We confirmed that the official release date of Hogwarts Legacy was February 10, 2023, which explains why the related issue emerged in February. \\
\hline
Sport Video &  Hello I was just wondering if I subscribe to MLB tv through Amazon Prime will i have access to any MLB Network shows Or just the games & Customers have raised questions about watching MLB games on Prime Video but have encountered technical issues. & We confirmed that the 2023 MLB season commenced in late February, which explains why the related issues emerged during that month. \\
\hline
TV Stick &  The screen says Amazon system recovery your Amazon fire tv will restart in a few minutes and should resume normal operation if it doesn't restart &  Customers have reported that their Fire TV devices received updates but failed to restart properly. & We checked Amazon's Digital \& Device Forum and confirmed that this issue has been widely discussed, and Amazon's engineering team has addressed it in the next update. \\
\hline
\hline
\end{tabular}
\caption{The table presents a selection of emerging issues identified in February 2023, categorized by product line. The first column displays the corresponding question sentences that represent the cluster center node. In the second column, we summarize the topic of each cluster. The third column shows the evidence that confirms these issues as emerging. It is noteworthy that the majority of these emerging issues are consistent with various online news sources.}
\label{table2}
\end{table*}
\end{document}